\documentclass[runningheads]{llncs}
\usepackage{graphicx}
\usepackage[table]{xcolor}
\definecolor{mygray}{gray}{.92}
\usepackage{tikz}
\usepackage{comment}
\usepackage{amsmath,amssymb} 
\usepackage{color}
\usepackage{booktabs}
\usepackage{multirow}  
\usepackage{bbding}
\usepackage{float}
\usepackage{subfig}
\usepackage{setspace}
\usepackage[symbol]{footmisc}

\makeatletter
\newcommand{\thickhline}{%
    \noalign {\ifnum 0=`}\fi \hrule height 1pt
    \futurelet \reserved@a \@xhline
}

\usepackage[accsupp]{axessibility}  

\usepackage[colorlinks,linkcolor=red,citecolor=green]{hyperref}
\begin{document}
\pagestyle{headings}
\mainmatter
\def\ECCVSubNumber{3187}  

\title{Domain Adaptive Person Search} 

\titlerunning{Domain Adaptive Person Search} 
\authorrunning{J. Li et al.} 
\author{Junjie Li$^{1,2}$\thanks{This work was done during Li's internship at Tencent Youtu Lab.}, Yichao Yan$^{1}$\thanks{Corresponding author: Yichao Yan.}, Guanshuo Wang$^2$, Fufu Yu$^2$,\\ Qiong Jia$^2$, Shouhong Ding$^2$}
\institute{$^1$MoE Key Lab of Artificial Intelligence, AI Institute, Shanghai Jiao Tong University \quad $^2$Tencent Youtu Lab\\
{\tt\small serenitycapo@gmail.com, yanyichao@sjtu.edu.cn, \\ 
\{mediswang,fufuyu,boajia,ericshding\}@tencent.com
}}

\maketitle

\begin{abstract}
Person search is a challenging task which aims to achieve joint pedestrian detection and person re-identification (ReID). Previous works have made significant advances under fully and weakly supervised settings. However, existing methods ignore the generalization ability of the person search models. In this paper, we take a further step and present Domain Adaptive Person Search (DAPS), which aims to generalize the model from a labeled source domain to the unlabeled target domain. Two major challenges arises under this new setting: one is how to simultaneously solve the domain misalignment issue for both detection and ReID tasks, and the other is how to train the ReID subtask without reliable detection results on the target domain. To address these challenges, we propose a strong baseline framework with two dedicated designs.
1) We design a domain alignment module including image-level and task-sensitive instance-level alignments, to minimize the domain discrepancy. 2) We take full advantage of the unlabeled data with a dynamic clustering strategy, and employ pseudo bounding boxes to support ReID and detection training on the target domain.
With the above designs, our framework achieves 34.7\% in mAP and 80.6\% in top-1 on PRW dataset, surpassing the direct transferring baseline by a large margin. Surprisingly, the performance of our unsupervised DAPS model even surpasses some of the fully and weakly supervised methods. The code is available at \url{https://github.com/caposerenity/DAPS}.
\end{abstract}


\section{Introduction}

Person search~\cite{DBLP:conf/cvpr/ZhengZSCYT17,DBLP:conf/cvpr/XiaoLWLW17} aims to detect and identify the query person from natural images. The mainstream approaches to tacking this task is to simultaneously address both tasks in an end-to-end manner, where supervised learning~\cite{DBLP:conf/cvpr/ZhengZSCYT17,DBLP:conf/cvpr/WangMCSC20,DBLP:conf/cvpr/ChenZYS20,DBLP:conf/aaai/LiM21} that rely on both pedestrian bounding boxes annotation and identity labels have been actively investigated. However, these supervised methods may suffer from significant performance degradation on unseen domains due to domain gaps.

\begin{figure}[t]
\setlength{\abovecaptionskip}{8pt}
\centering
\includegraphics[width=\linewidth]{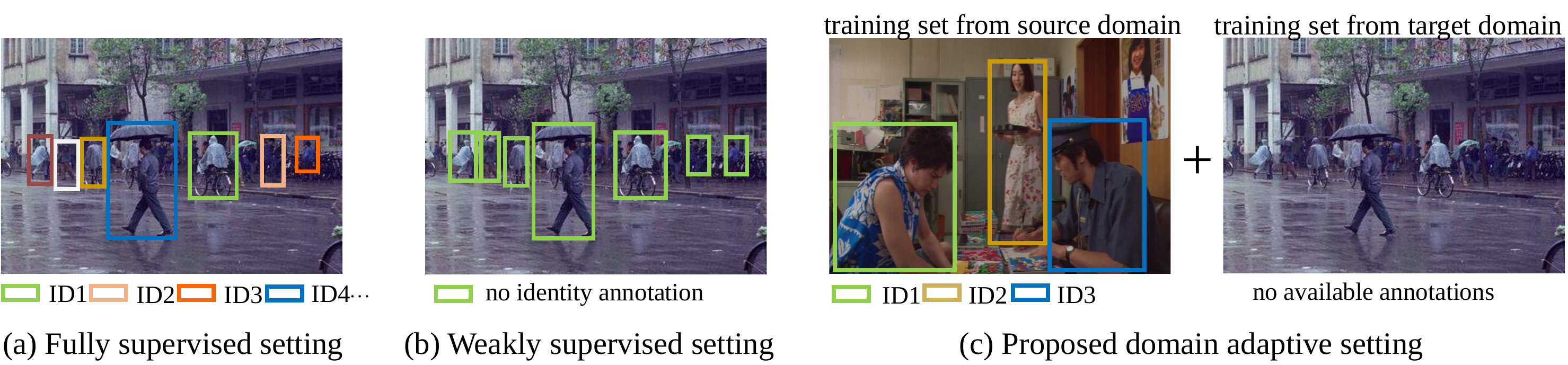}
\caption{Comparison of three person search settings. (a) Fully supervised setting: bounding boxes and identity annotations are available. (b) Weakly supervised setting: only bounding boxes annotations are available. (c) Domain adaptive setting: neither bounding boxes nor identity annotations on the target domain is accessible, and there exists obvious domain gaps between different domains, \emph{e.g.}, the size of human crops. The network is trained with both the labeled source domain and the unlabeled target domain images.}
 \label{fig:intro}
\end{figure}

To address this problem, several recent works~\cite{yan2022exploring,han2021weakly} propose the weakly supervised person search (WSPS) setting without accessible ID annotations, shown in Fig.~\ref{fig:intro}. Nevertheless, several limitations are still waiting to be addressed. First, these works still require manual annotation of the ground-truth bounding boxes for the detection task, which obviously is not an economical option for real-world applications. Second, there exist several large-scale annotated person search datasets, e.g., CUHK-SYSU~\cite{DBLP:conf/cvpr/XiaoLWLW17} and PRW~\cite{DBLP:conf/cvpr/ZhengZSCYT17}, 
which can serve as supervised source domains and help improve the performance on the unlabeled target data. Unfortunately, the weakly supervised setting does not fully unleash the potential of the available training data. Third, these methods adopt an inconsistent training strategy with supervised detection and unsupervised ReID, which ignores the essential correlation between the two sub-tasks.

Inspired by the unsupervised domain adaptation (UDA)~\cite{DBLP:conf/icml/GaninL15,DBLP:conf/cvpr/Kang0YH19,DBLP:journals/ijon/WangD18}, as shown in Fig.~\ref{fig:intro}, we present the Domain Adaptive Person Search (DAPS) framework, where person search models trained on labeled source domain are transferred to unlabeled target domains. Compared to weakly supervised person search, neither the identity labels nor the bounding boxes are accessible in DAPS. Our framework faces two major challenges: (1) Both the detection and the ReID sub-tasks suffer from domain gap. However, detection focuses on the commonness of people regardless of the identities, while ReID needs to learn the uniqueness of different persons. This conflict can be more serious in domain adaptation. (2) Since the ground-truth detection boxes are not available, it will be extremely challenging to accurately localize the pedestrians in the target domain, which further increases the difficulty for the ReID sub-task. Therefore, directly extending WSPS methods to take advantage of target domain data is infeasible.

To address the first challenge, we explore domain alignment for robust domain invariant feature learning. In the context of pedestrian detection, this is typically achieved by domain adversarial training \cite{DBLP:conf/cvpr/Chen0SDG18} on both image-level and instance-level features. 
Following this line of research, we design a domain alignment module (DAM) to alleviate the discrepancy between different domains. Specifically, on the one hand, we introduce domain discriminators at intermediate backbone layers. On the other hand, we perform a task-sensitive instance-level alignment to mitigate the conflicts between two sub-tasks. We observe that such a domain alignment operation is beneficial for both branches.

To tackle the second challenge, 
we generate pseudo bounding boxes on the target domain images iteratively, and perform the training process with GT and pseudo boxes for domain adaptation. Furthermore, we present a dynamic clustering strategy to generate pseudo identity labels on the target domain. To fully release the potential of the target domain training data, the proposed framework refines the detection task with selected proposals, and enhances the interaction between the two sub-tasks with hybrid hard case mining. Experimental results demonstrate that this design surprisingly achieves comparative performance with directly adopting ground-truth bounding boxes.

Our contributions are summarized as three-fold:
\begin{itemize}
    \setlength{\itemsep}{2pt}
	\setlength{\parsep}{-3pt}
	\setlength{\parskip}{-0pt}
	\setlength{\leftmargin}{-15pt}
    \item We introduce a novel unsupervised domain adaptation paradigm for person search. This setting requires neither bounding boxes nor identity annotations on the target domain, making it more practical for real-world applications. 

    \item We present the DAPS framework to overcome the challenges caused by cross-domain discrepancy and cross-task dependency. We propose domain alignment for person search to enhance domain-invariant feature learning. Meanwhile, a dynamic clustering and a hybrid hard case mining strategy are introduced to facilitate unsupervised target domain learning.

    \item Without any auxiliary label in the target domain, our framework achieves promising performance on two target person search benchmarks, surprisingly outperforming several weakly and fully supervised models.
\end{itemize}

\section{Related Work}

\subsection{Person Search}
With the development of deep learning and large scale benchmarks~\cite{DBLP:conf/cvpr/XiaoLWLW17,DBLP:conf/cvpr/ZhengZSCYT17}, person search~\cite{DBLP:conf/aaai/ChenZO0S20} has recently become a popular research topic. Existing fully supervised person search models can be divided into two-step and one-step frameworks. Two-step frameworks typically consist of separately trained detection and ReID models~\cite{DBLP:conf/cvpr/WangMCSC20,DBLP:conf/iccv/HanYZTZGS19}. Zheng \emph{et al.}~\cite{DBLP:conf/cvpr/ZhengZSCYT17} make a systematic evaluation on different combination of detection and ReID models. Wang \emph{et al.}~\cite{DBLP:conf/cvpr/WangMCSC20} solve the inconsistency between detection and person ReID tasks. One-step frameworks~\cite{DBLP:conf/cvpr/ChenZYS20,DBLP:conf/aaai/LiM21,DBLP:conf/cvpr/YanLQBL00021} design a unified model to jointly solve detection and ReID tasks in an end-to-end manner, making the pipeline more efficient. Yan \emph{et al.}~\cite{DBLP:conf/cvpr/YanZNZXY19} introduce a graph model to explore the impact of contextual information for identity matching. Chen \emph{et al.}~\cite{DBLP:conf/cvpr/ChenZYS20} disentangle the person representation into norm and angle to eliminate the cross-task conflict. Li \emph{et al.}~\cite{DBLP:conf/aaai/LiM21} develop a sequential structure to reduce the low-quality proposals. Several recently studies~\cite{yan2022exploring,han2021weakly} adopt the weakly supervised setting without no accessible person ID labels. In this work, we explore a novel person search setting to generalize labeled source to unlabeled target domain without any bounding boxes and ID labels annotation.

\subsection{Domain Adaptation for Person ReID}
Unsupervised domain adaptation (UDA) ReID~\cite{DBLP:conf/iccv/ChenZG19,DBLP:conf/cvpr/Deng0YK0J18,DBLP:conf/cvpr/LiuCS20,DBLP:conf/iccv/FuWWZSUH19,DBLP:conf/nips/Ge0C0L20,DBLP:journals/pr/SongWZDZHW20,DBLP:journals/corr/abs-2111-14290} typically trains a model with labelled source domain and transfers to the target domain under the unsupervised setting. Mainstream UDA ReID methods can be divided into two categories. The first category employs generative adversarial networks~\cite{DBLP:conf/nips/GoodfellowPMXWOCB14} to mitigate the style discrepancy and translate the labelled source domain data into the target domain~\cite{DBLP:conf/iccv/ChenZG19,DBLP:conf/cvpr/Deng0YK0J18,DBLP:conf/cvpr/LiuCS20}. For the second category, they generate pseudo labels by clustering~\cite{DBLP:conf/iccv/FuWWZSUH19,DBLP:conf/nips/Ge0C0L20,DBLP:journals/pr/SongWZDZHW20} or assigning soft labels~\cite{DBLP:conf/cvpr/WangZ20a} on target domain, and use these pseudo labels to further supervise target domain training. Recently, pseudo label-based methods raise more attention due to their superior  performance. However, UDA ReID requires the cropped images, which cannot be directly extended to adaptive person search due to the lack of bounding boxes on target domain. To address this, we propose a dynamic clustering strategy to generate high-quality pseudo boxes to facilitate target domain training.

\subsection{Domain Adaptive Object Detection}
Existing Domain Adaptive approaches object detection can be categorized into three main branches, including adversarial-based methods~\cite{DBLP:conf/cvpr/TzengHSD17,DBLP:conf/cvpr/Chen0SDG18,DBLP:conf/cvpr/ZhuPYSL19,DBLP:conf/cvpr/WangZYF19,DBLP:conf/cvpr/SaitoUHS19}, discrepancy-based methods~\cite{DBLP:conf/iccv/KhodabandehVRM19,DBLP:conf/cvpr/CaiPNTDY19,DBLP:journals/inffus/CaoGHYCQ19} and reconstruction-based methods~\cite{DBLP:conf/icip/Lin19,DBLP:conf/ijcnn/ArrudaPBSBSO19,DBLP:conf/cvpr/DevaguptapuASB19}. Adversarial-based methods utilize a domain discriminator to distinguish the domain of input data, the adversarial training is performed to encourage domain confusion between the source domain and the target domain. The discrepancy-based strategy utilizes the unlabeled target domain images to fine-tune the detector, and further followed by mean-teacher learning~\cite{DBLP:conf/cvpr/CaiPNTDY19} or auto-annotation~\cite{DBLP:journals/inffus/CaoGHYCQ19}. The reconstruction-based approaches bridge the domain gap by reconstructing the source or target samples, which is usually realized by image-to-image translation~\cite{DBLP:conf/ijcnn/ArrudaPBSBSO19,DBLP:conf/icip/Lin19}. In this work, we consider the conflicts between sub-tasks of person search, and develop a task-sensitive alignment module to alleviate such conflicts.

\begin{figure}[t]
    \setlength{\abovecaptionskip}{8pt}
    \centering
    \includegraphics[width=\linewidth]{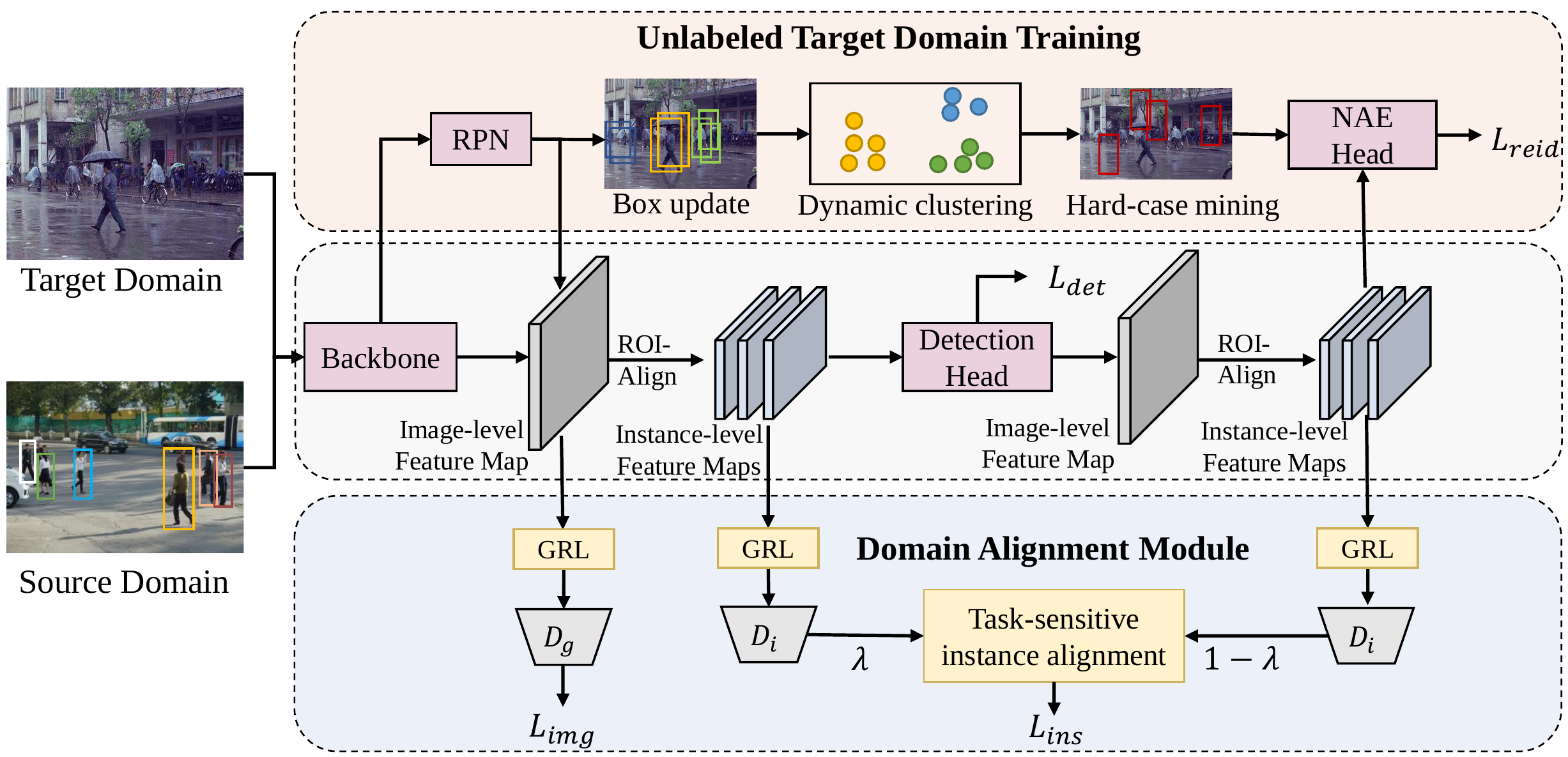}
    \caption{Architecture of the DAPS framework. ``GRL'' denotes the gradient reverse layer~\cite{DBLP:conf/icml/GaninL15}. The backbone follows SeqNet~\cite{DBLP:conf/aaai/LiM21}, and we employ a domain alignment module to minimize domain discrepancy on both image-level and instance-level. We further impose dynamic clustering, hybrid hard case mining and target detection training to take full advantage of the unlabeled target domain data.}
    \label{fig:pipline1}
\end{figure}

\section{Methodology}

\subsection{Framework Overview}
The general pipeline of the proposed DAPS framework is illustrated in Fig.~\ref{fig:pipline1}. 
Given the input images from both the source and the target domain, the image-level feature maps are extracted with a backbone network. Then, these features are input into the Region Proposal Network (RPN) to generate candidate bounding boxes, which are subsequently fed into the ROI-Align layer to represent instance-level feature maps. To close the domain gaps for the downstream detection and ReID tasks, we design a domain alignment module (DAM) to align both image-level and instance-level features from different domains. 

Subsequently, the domain-aligned instance-level feature maps are input into both the detection and the ReID branch. Since the ground-truth bounding boxes are not available in the target domain, the model will generate different pedestrian detection results for each training epoch. Therefore, it is infeasible to follow the traditional UDA ReID methods, which generally perform clustering on a fixed size of instances to generate pseudo labels. To address this issue, we design a novel dynamic clustering strategy, which continuously associates the bounding boxes generated from consecutive  epochs, to guarantee the stability of instance-level ReID features. Based on dynamic clustering strategy, we further introduce the hybrid hard case mining and the target domain detection refinement to sufficiently take advantage of the unlabeled training data.

\subsection{Domain Alignment Module}
\textbf{Image-level Alignment.}
As discussed in~\cite{DBLP:journals/ijon/WangD18,DBLP:conf/cvpr/Chen0SDG18,DBLP:conf/iccv/ChenZG19,DBLP:conf/cvpr/Deng0YK0J18}, minimizing domain discrepancy is beneficial for both sub-tasks of person search, and an effective way is to guide the model to learn domain-invariant representation. Motivated by the recent progress in domain adaptive detectors \cite{DBLP:conf/cvpr/Chen0SDG18,DBLP:conf/cvpr/ZhuPYSL19,DBLP:conf/cvpr/WangZYF19,DBLP:conf/cvpr/SaitoUHS19}, where intermediate features are imposed with image-level alignment constraints, we introduce a domain alignment module into our DAPS framework. As shown in Fig.~\ref{fig:pipline1}, DAM employs a patch-based domain classifier to predict the domain where the input feature comes from. A min-max formulation is adopted to misdirect the domain classifier and encourage domain-invariant representation learning.

Suppose we have $N$ training images  $\{{I}_1, ..., {I}_N\}$ with corresponding domain labels $\{{d}_1, ..., {d}_N\}$. Particularly, ${d}_i = 0$ indicates that image ${I}_i$ comes from the source domain, while ${d}_i = 1$ denotes the target domain. We denote the backbone of DAPS as $\mathrm\Phi$ and the image-level domain classifier as $\mathrm{D}_g$, and further represent the domain prediction result of input ${I}_i$ as ${p}_i$.
We apply a cross entropy loss to perform domain alignment in an adversarial training manner:
\begin{equation}
    \mathcal{L}_{i m g}=-\sum_{i}\left[{d}_{i} \log {p}_{i}+\left(1-{d}_{i}\right) \log \left(1-{p}_{i}\right)\right].
\end{equation}
We have tried to conduct image-level alignment on different intermediate features and multi-scale alignment, but achieve no better results.

\begin{figure}[t]
  \setlength{\abovecaptionskip}{8pt}
  \centering
  \includegraphics[width=\linewidth]{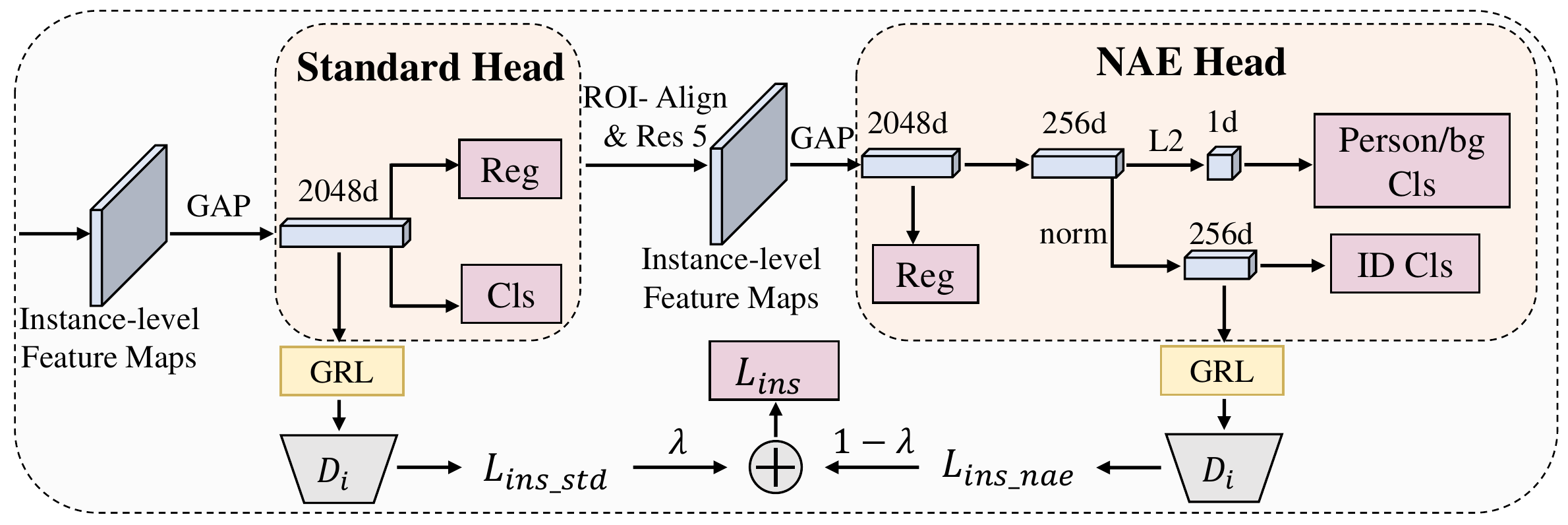}
   \caption{Details of the two heads and the task-sensitive instance-level alignment.}
   \label{fig:r1}
\end{figure}
\textbf{Task-sensitive Instance-level Alignment}.
As illustrated in Fig.~\ref{fig:r1}, our framework consists of two head networks, where the detection performance mainly depends on the first standard Faster R-CNN \cite{DBLP:conf/cvpr/GirshickIDM15} head, while the NAE \cite{DBLP:conf/cvpr/ChenZYS20} head is highly relevant to ReID. When the scale of the source domain is much smaller than the unlabeled target, the target pseudo bounding boxes predicted by detector trained on the source can be severely overfitted to the smaller domain, but no reliable target detection guidance can relieve this issue. When the target is much smaller, pseudo target ID labels can be easily obtained by clustering, but these might provide insufficient generalizing for the ReID sub-task. 

According to the characteristics of the up- and down-stream tasks, we propose the task-sensitive instance-level alignment module by balancing the alignment weight on instance-level features for both sub-tasks.
Suppose we have ${K}_1$ instances in the standard head and ${K}_2$ instances for the NAE head, two domain classifiers $\{{D}_i^{d}, {D}_i^{r}\}$ are built in the same way with image-level alignment, and the domain predictions of the two local classifiers are denoted as $\{{p}_{i,1}^{d}, ..., {p}_{i,K_1}^{d}\}$, $\{{p}_{i,1}^{r}, ..., {p}_{i,K_2}^{r}\}$, respectively. The instance-level loss can be formulated as:
\begin{equation}
\begin{aligned}
    \mathcal{L}_{i n s}= &- \lambda \sum_{i, j}\left[{d}_{i} \log {p}_{i,j}^{d}+\left(1-{d}_{i}\right) \log \left(1-{p}_{i,j}^{d}\right)\right]    \\
    &- (1-\lambda) \sum_{i, k}\left[{d}_{i} \log 
    {p}_{i,k}^{r}+\left(1-{d}_{i}\right) \log \left(1-{p}_{i,k}^{r}\right)\right].
\end{aligned}
\end{equation}
where $j \in \{1,...,{K}_1\}$, and $k \in \{1,...,{K}_2\}$. The source and target domain contains $N_s$ and $N_t$ images respectively, and the balancing factor $\lambda$ is obtained by
\begin{equation}
    \lambda = \sigma\left(4\cdot  {sign}\left(N_t - N_s \right)
    \left(\frac{\max(N_s, N_t)}{\min(N_s, N_t)}-1\right) \right).
\end{equation}
where the $\sigma\left( \cdot \right)$ is Sigmoid function to normalize the domain scale ratio. Moreover, we impose a L2-norm regularizer to ensure the consistency between image-level and instance-level classifiers.

\subsection{Training on Unlabeled Target Domain}
\textbf{Dynamic Clustering}. UDA ReID models typically employ the clustering strategy (e.g., DBSCAN) to generate pseudo labels for the target domain instances, and employ memory-based losses~\cite{DBLP:conf/nips/Ge0C0L20} for metric learning. However, without ground-truth bounding boxes on target domain, the instances can be only generated from the detection results, which varies with the training process. This makes it infeasible to directly apply typical clustering approach to DAPS. To address this issue, we propose a novel dynamic clustering strategy to make full use of the detection results for continuous ReID training.

\begin{figure}[t]
    \setlength{\abovecaptionskip}{8pt}
    \centering
    \includegraphics[width=\linewidth]{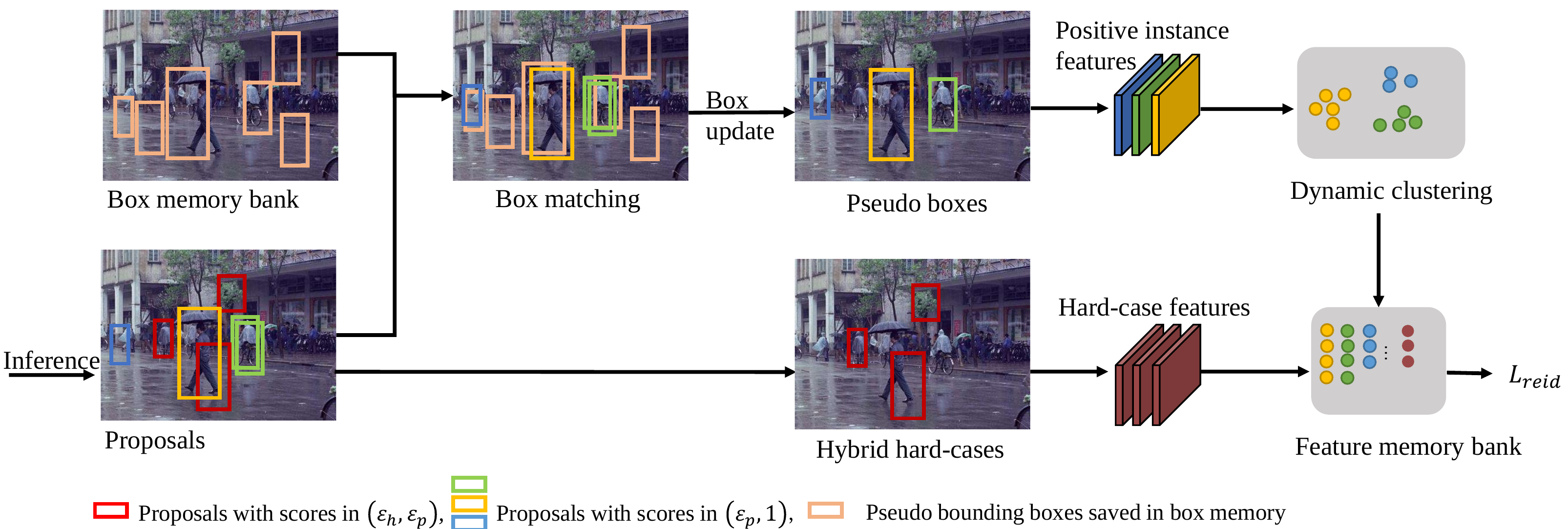}
    \caption{Illustration of the dynamic clustering and hard case mining. At start of each epoch, we employ generated proposals, including both qualified ones and hard cases, to update the memory bank. Qualified proposals are adopted for matching pseudo boxes memory, and hard cases will directly be added.}
    \label{fig:dynamic}
\end{figure}


As illustrated in Fig.~\ref{fig:dynamic}, an asynchronized training strategy is introduced to progressively update pseudo bounding boxes with the selected proposals as ground-truth boxes on the target domain. Specifically, for the beginning $\alpha$ epochs, DAPS is trained only on the source dataset labeled with both bounding boxes and ID labels. After that, we maintain a bounding box memory $\mathbf {M_B} = \{{B}_1,...,{B}_{N_t}\}$ and a feature vector memory $\mathbf {M_V} = \{{V}_1,...,{V}_{N_t}\}$, corresponding to each of $N_t$ target domain images. 
At the start of each subsequent epochs, DAPS filters out high-confidence candidate proposals $\{{c}_1,...,{c}_{m}\}$ from $x_{i}^{t}$, and employ them to match pseudo bounding boxes in box memory $B_i  = \{{b}_1,...,{b}_{n}\}$ according to IOU scores. Every proposal is assigned to the most relevant box in memory if their IOU score is above the threshold, and the boxes which fail to match any qualified proposal will be removed from memory $B_i$. The remaining boxes in the memory are continuously updated in the Exponential Moving Average (EMA) method.

For example, suppose the proposals $c_{j1}, c_{j2}, c_{j3}$ are mapped to the box $b_k$, then $b_k$ is updated by:
\begin{equation}
    b_k \leftarrow \gamma b_k+(1-\gamma) \mathrm{avg} \left(c_{j1}, c_{j2}, c_{j3}\right),
\end{equation}
where $\gamma \in \left[0,1 \right]$ controls the update rate. Eventually, the proposals without any matched box will also be fed into the memory $B_i$, and further, the feature memory $\mathbf {M_V}$ is updated in the same way. Afterwards, we perform clustering upon $\mathbf {M_V}$ to obtain $N_{t}^{c}$ clusters $\{{C}_1,...,{C}_{N_{t}^{c}}\}$ with centroids $\mathbf {W} = \{{w}_1,...,{w}_{N_{t}^{c}}\}$, and $N_{t}^{o}$ instances $\mathbf {F} = \{{f}_1,...\, {f}_{N_{t}^{o}}\}$ not belonging to any cluster. By extracting the identity features $\mathbf {V}$ in the source domain, we eventually build a unified memory $\mathbf {M} = \{\mathbf {V} ,\mathbf {W},\mathbf {F}\}$ for ReID training. The loss function can be expressed :
\begin{equation}\label{eq:memory_loss}
\begin{aligned}
\mathcal{L}_{} & =-\log \frac{\exp \left({x} \cdot {z}^{+} / \tau\right)}
{\sum_{k=1}^{N_{t}^{c}} \exp \left({x}\cdot {w}_{k} / \tau\right)+
\sum_{k=1}^{N_{t}^{o}} \exp \left({x}\cdot {f}_{k} / \tau\right)+
\sum_{k=1}^{N_{s}^{c}} \exp \left({x}\cdot {v}_{k} / \tau\right)},
\end{aligned}
\end{equation}
where ${w}$, ${f}$, and ${v}$ represents the target domain clusters, the independent instances and the source domain classes, respectively. ${z}^{+}$ is the corresponding class prototype of the input feature $x$, and $\cdot$ denotes the inner product to measure the feature similarity. The features in the memory will be updated in a momentum way during backward stage:
\begin{equation}
    z_t \leftarrow \gamma z_t+(1-\gamma)x,
\end{equation}
where $z_t$ is the $t$-th prototype in the memory bank $\mathbf M$.

\textbf{Hybrid Hard Case Mining}.
A significant challenge for dynamic clustering is to generate reliable bounding boxes. We treat those boxes with lower confidence than a threshold as negative samples. 
In order to sufficiently exploit target domain information, we explore the potential of adding these ``negative'' samples to the ReID training.
Proposals with relatively low confidence scores can be divided into highly overlapped with high-confidence boxes, the undetected persons and the background clutters. It is undesirable to enhance the ReID sub-task by treating all these proposals as negative samples. As a result, we design a hierarchical scheme to categorize the candidate proposals, and employ both of the low-confidence person proposals and non-trivial background clutters to enhance the discrimination of the ReID branch. 

Specifically, proposals with confidence score in the range of $\left(\epsilon_h, \epsilon_p \right)$ defined by upper and lower bound thresholds are regarded as non-trivial cases. We exclude highly overlapped duplicates by further screening IOUs with positive proposals, while the hybrid of undetected persons and the negative clutters are reserved for training. The features of these hard cases will be added to $\mathbf{M}$, and be used for the contrastive learning process. The memory loss in Eq.~\ref{eq:memory_loss} is modified as:
\begin{equation}
\begin{aligned}
\mathcal{L}_{}=&-\log \frac{\exp \left({x} \cdot {z}^{+} / \tau\right)} {\sum_{z \in \mathbf{M}}{\exp \left( {x \cdot z/\tau} \right)}}, \\
\sum_{z \in \mathbf{M}}{\exp \left( {x \cdot z/\tau} \right)} = &\sum_{k=1}^{N_{t}^{c}} \exp \left({x}\cdot {w}_{k} / \tau\right)+
\sum_{k=1}^{N_{t}^{o}} \exp \left({x}\cdot {f}_{k} / \tau\right)+ \\
&\sum_{k=1}^{N_{s}^{c}} \exp \left({x}\cdot {v}_{k} / \tau\right)+
\sum_{k=1}^{N_{t}^{n}} \exp \left({x}\cdot {h}_{k} / \tau\right),
\end{aligned}
\end{equation}
where $h$ denotes the hybrid hard cases.
It is noteworthy that the hybrid hard cases will be involved into the dynamic clustering before the next epoch. Once a hard case is matched with new qualified proposals, it will be treated as a positive sample and updated in a momentum way.

\textbf{Target Detection Training}.
Although DAM can minimize the domain discrepancy, the over-fitting towards the source domain is still likely to take place, especially when the source domain data is extremely less complex and comprehensive than the target domain images. To this end, simultaneously training detection with both of the source and the target domain data is beneficial for the generalization ability of model. DAM and dynamic clustering provide relatively reliable pseudo bounding boxes, and specifically, we employ such pseudo bounding boxes after the $\alpha$ epoch to supervise detection on the target domain. In this way, the potential of unlabeled target domain images is released for both ReID and detection training.

\section{Experiment}


\subsection{Datasets and Evaluation Protocols}
\textbf{Datasets.}\label{dataset}
We employ two large-scale benchmark datasets, CUHK-SYSU~\cite{DBLP:conf/cvpr/XiaoLWLW17} and PRW~\cite{DBLP:conf/cvpr/ZhengZSCYT17} in our experiments. CUHK-SYSU is one of the largest public datasets for person search, composed of 18,184 images and 96,143 bounding boxes from 8,432 different identities. It is divided into a training set of 11,206 images with 5,532 identities, and a test set with 6,978 gallery images and 2,900 query images. The widely used PRW dataset contains 11,816 images, 43,110 annotated bounding boxes from 932 identities. The training set includes 5,704 images and 482 labelled persons, while the other 6,112 images and 2,057 probe persons from 450 identities are adopted as test set. 

\textbf{Evaluation Protocols.}
Our experiments employ the default splits for both datasets. For domain adaptation settings, the annotations of dataset used as the source domain is accessible, while neither bounding boxes nor identity labels of datasets as the target domain are available. All evaluations are performed on the test set of target domain. We adopt the widely used mean average precision (mAP) and cumulative matching characteristic (CMC) top-1 accuracy as evaluation metrics for ReID sub-task, while average precision (AP) and recall rate are adopted as the metrics for detection.

\subsection{Implementation Details}
We adopt ResNet50~\cite{DBLP:conf/cvpr/HeZRS16} pretrained on ImageNet-1k~\cite{DBLP:conf/cvpr/DengDSLL009} as our default backbone network. DBSCAN~\cite{DBLP:conf/kdd/EsterKSX96} with self-paced learning strategy~\cite{DBLP:conf/nips/KumarPK10} is employed as the basic clustering method, we set default hyper-parameters $\epsilon_p = 0.95$, $\epsilon_h = 0.8$ and $\lambda_t = 0.1$. During training, the input images are resized to $1500\times900$, and random horizontal flip is applied for data augmentation. Our model is optimized by Stochastic Gradient Descent (SGD) for 20 epochs. We set a mini-batch size of 4, and an initial learning rate of 0.0024, which is reduced by a factor of 0.1 at epoch 16 with warmed up in the first epoch. The momentum and weight decay are set to 0.9 and $5 \times 10^{-4}$, respectively. We set the momentum factor $\gamma$ for memory updating to 0.2. The starting epoch of $\alpha$ is set to 8 when PRW is chosen as target domain, and 0 for CUHK-SYSU. All experiments are implemented with one NVIDIA Tesla A100 GPU. We also plan to support this project with MindSpore in our future work.


\begin{table}[t]
\setlength{\abovecaptionskip}{2mm}
\centering
\caption{Comparative results when combining different components. DAM: Domain Alignment Module. DC: Dynamic Clustering. HM: Hybrid hard case Mining. DTD: Detection on Target Domain.}
\begin{tabular}{cccc|cccc|cccc}
\hline\thickhline
\rowcolor{mygray}  
              &     &     &     & \multicolumn{4}{c|}{Target: PRW}    & \multicolumn{4}{c}{Target: CUHK-SYSU}                        \\ \cline{5-12} 
\rowcolor{mygray}  
 {\multirow{-2}{*}{DAM}}                  &  {\multirow{-2}{*}{DC}}            &  {\multirow{-2}{*}{HM}}  &  {\multirow{-2}{*}{DTD}}      & mAP     & top-1  & recall           & \multicolumn{1}{c|}{AP}                  & mAP     & top-1  & recall           & \multicolumn{1}{c}{AP} \\ 
\hline \hline
$\times$   &$\times$  & $\times$ & $\times$ & 30.3  & 77.7     & 94.0  & 88.3  & 52.5  & 54.8  & 55.2 &55.1  \\
$\checkmark$    & $\times$ & $\times$ & $\times$ &30.9  &79.3  &96.3  &90.7   &62.2  &63.6  &70.8  &63.1                \\
$\times$    & $\checkmark$ & $\times$ & $\times$ &32.2  &79.4  &96.8  &90.3 &70.9  &72.3  &67.8 &62.2        \\
$\checkmark$  &$\checkmark$ & $\times$ & $\times$ &32.7  &79.6  &95.9  &90.4  &72.6  &74.3  &68.3  &63.2               \\ \hline
$\checkmark$  & $\checkmark$ & $\checkmark$ & $\times$  &{34.5}   & \textbf{80.7} & 97.0 & 91.0 &73.2   & 74.8 & 70.4 & 64.1                \\
$\checkmark$    & $\checkmark$ & $\times$ & $\checkmark$ &33.1  &79.9  &96.6  &\textbf{91.2} &76.8  &78.7  &\textbf{79.4}  &\textbf{71.1}         \\
$\checkmark$  & $\checkmark$ & $\checkmark$ &$\checkmark$ &\textbf{34.7}  &80.6  &\textbf{97.2}  &90.9  &\textbf{77.6}  &\textbf{79.6}  &77.7 &69.9 \\ \hline
\end{tabular}
\label{tab:ablation}
\end{table}

\begin{table}[t]
\setlength{\abovecaptionskip}{2mm}
\centering
\caption{Comparative results of task-sensitive instance-level alignment.}
\begin{tabular}{c|cccc|cccc}
\hline\thickhline
\rowcolor{mygray}  
              &     \multicolumn{4}{c|}{Target: PRW}    & \multicolumn{4}{c}{Target: CUHK-SYSU}                        \\ \cline{2-9} 
\rowcolor{mygray}  
 {\multirow{-2}{*}{instance da}}      & mAP     & top-1  & recall           & \multicolumn{1}{c|}{AP}                  & mAP     & top-1  & recall           & \multicolumn{1}{c}{AP} \\ 
\hline \hline  
normal &21.7  &76.0  &\textbf{96.7}  &\textbf{91.1}  &58.2  &60.5  &66.3  &56.3 \\
task-sensitive   & \textbf{30.9} & \textbf{79.3}   &96.3 & 90.7 & \textbf{62.2} & \textbf{63.6}    & \textbf{70.8} & \textbf{63.1}\\ \hline
\end{tabular}
\label{tab:task-sensitive}
\end{table}

\begin{table}[t]
\setlength{\abovecaptionskip}{2mm}
\centering
\caption{Comparative results when employing different strategies to handle the lack of bounding boxes. `GT' refers to using the ground truth bounding boxes for all the training process of ReID, and `GT for init' only employs these boxes to initialize the memory bank. `static' means directly employing the qualified proposals before each epoch.}
\begin{tabular}{c|cccc|cccc}
\hline\thickhline
\rowcolor{mygray}  
              &     \multicolumn{4}{c|}{Target: PRW}    & \multicolumn{4}{c}{Target: CUHK-SYSU}                        \\ \cline{2-9} 
\rowcolor{mygray}  
 {\multirow{-2}{*}{strategy}}      & mAP     & top-1  & recall           & \multicolumn{1}{c|}{AP}                  & mAP     & top-1  & recall           & \multicolumn{1}{c}{AP} \\ 
\hline \hline  
GT & 34.9  &79.9  &94.9  &89.5  &73.6  &76.0  &74.6  &68.2 \\
GT for init &33.5  &79.6  &92.9  &88.5 &73.5  &75.4  &64.4  &60.8 \\ \hline
Static   &25.3  &77.3  &96.6  &90.8 &64.0  &66.1  &67.6  &62.5    \\         
Dynamic Update  & 32.7 & 79.6    & 95.9 & 90.4 &72.6  &74.3  &68.3  &63.2\\ \hline
\end{tabular}
\label{tab:memory}
\end{table}

\begin{table}[t]
\setlength{\abovecaptionskip}{2mm}
\centering
\caption{Comparative results of when to start asynchronized training.}
\begin{tabular}{c|cccc|cccc}
\hline\thickhline
\rowcolor{mygray}  
              &     \multicolumn{4}{c|}{Target: PRW}    & \multicolumn{4}{c}{Target: CUHK-SYSU}                        \\ \cline{2-9} 
\rowcolor{mygray}  
 {\multirow{-2}{*}{starting epoch}}      & mAP     & top-1  & recall           & \multicolumn{1}{c|}{AP}                  & mAP     & top-1  & recall           & \multicolumn{1}{c}{AP} \\ 
\hline \hline  
0 &31.5  &79.7  &95.8  &89.4  &\textbf{77.6}  &\textbf{79.6}  &\textbf{77.7}  &\textbf{69.9} \\
4 &31.4  &79.4  &95.8  &89.1 &73.6  &75.3  &76.6  &67.7 \\
8   & \textbf{34.7} & \textbf{80.6}    & 97.2 & \textbf{90.9} & 73.2 & 74.7    & 76.7 & 69.0\\
10 &33.4  &80.6  &\textbf{97.5}  &90.7  &71.4  &73.3  &74.8  &65.8\\ \hline
\end{tabular}
\label{tab:late launch}
\end{table}

\begin{figure}[t]
\setlength{\abovecaptionskip}{8pt}
\centering
\subfloat[\label{subfig-1-1}]{%
   \includegraphics[width=0.4\linewidth]{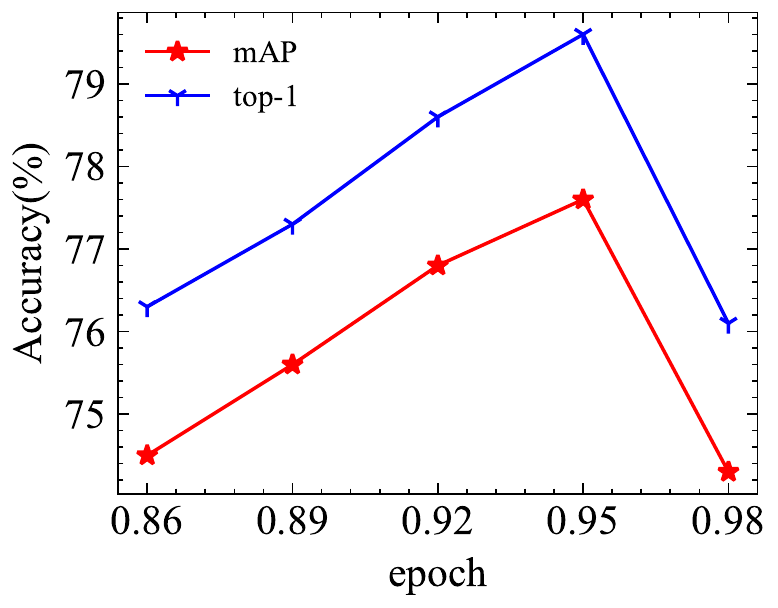}
}
\hspace{8mm}
\subfloat[\label{subfig-1-2}]{%
   \includegraphics[width=0.4\linewidth]{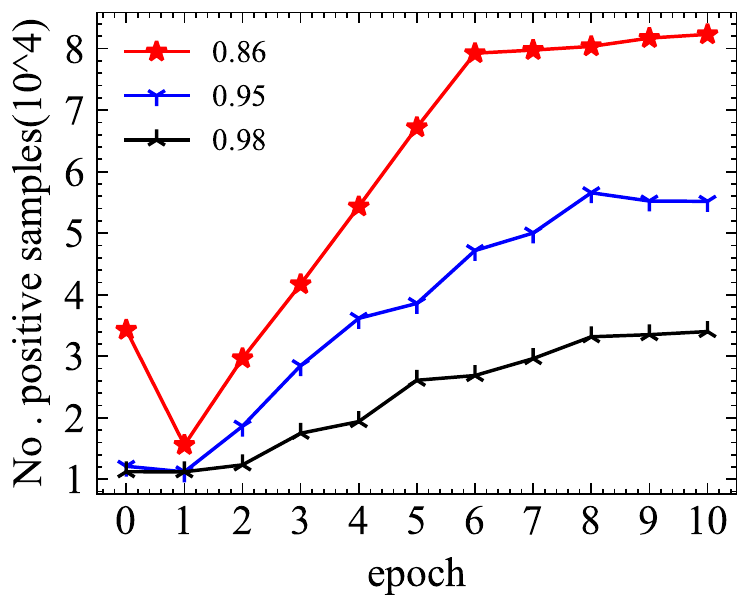}
}
\caption{Target domain performance with different $\epsilon_p$ on CUHK-SYSU dataset. (a): ReID accuracy results; (b): Numbers of generated positive proposals. The ground truth instances number is 55,260.}
 \label{fig:eps}
\end{figure}

\subsection{Ablation Study}
We perform analytical experiments to verify the effectiveness of each detailed component in our proposed framework. In table~\ref{tab:ablation}, we compare the baseline method with different combinations of proposed components, and report the results on both CUHK-SYSU and PRW datasets. For example, when we use CUHK-SYSU as the target domain dataset, the directly transferring baseline model achieves 52.5\% mAP and 54.8\% top-1. After individually adding the domain adaptive module (DAM) and dynamic clustering (DC), the performance improves 9.3\% and 18.4\% in terms of mAP. When combining DAM and DC, the mAP is further promoted to 72.6\%, surpassing the 52.5\% of baseline by a large margin. Furthermore, to make full use of the unlabeled target data, we implement the hybrid hard case mining (HM) and detection on target domain (DTD). HM improves the ReID performance by 0.6\% in mAP, and DTD prominently enhances the detection branch with a 7.0\% gain for AP. Eventually, DAPS achieves 77.6\% mAP and 79.6\%top-1 with all designed modules, outperforming the baseline by 25.1\% in mAP, 24.8\% in top-1, 22.5\% in recall, and 14.8\% in AP.

\textbf{Effectiveness of task-sensitive instance-level alignment}.
To validate the effectiveness of our task-sensitive instance-level alignment design, we compare it with normal instance-level alignment, which conducts instance alignment on both head networks without balancing between them. As observed in Table~\ref{tab:task-sensitive}, the task-sensitive design successfully alleviates the inner task conflicts and outperforms normal strategy by a large margin.

\textbf{Effectiveness of dynamic clustering}.
As aforementioned, the key to utilizing unlabeled target domain data is generating reliable pseudo bounding boxes. To validate the quality of the pseudo bounding boxes we use, we compare different strategies of obtaining bounding boxes, and the results are reported in Table~\ref{tab:memory}. We first measure the performance achieved by using ground-truth bounding boxes for training the ReID task. Furthermore, we report the performance achieved by directly employing the qualified proposals before each epoch, which is denoted as `static' in Table~\ref{tab:memory}. The results reveal that our proposed dynamic clustering strategy can generate trustworthy pseudo bounding boxes to achieve comparable performance with using ground-truth boxes.

\textbf{Effectiveness of asynchronized training}.
We conduct experiments for influences by the training stage hyper-parameter $\alpha$ on final performance. As shown in Table~\ref{tab:late launch}, when PRW is adopted as the target domain, the best performance is achieved with $\alpha=8$, while with $\alpha=0$ for CUHK-SYSU. The results might be counterintuitive but indeed validate our task-sensitive motivation. For smaller source dataset, even limited additional target information might be helpful for cross-domain generalization. In contrast, for larger source dataset, unreliable target proposals can be harmful for domain gap bridging.


\textbf{Analysis on hyper-parameter $\epsilon_p$}.
We visualize the influence of hyper-parameter $\epsilon_p$ in Fig.~\ref{fig:eps}. We observe that the value of $\epsilon_p$ influences the ReID performance to a large extend, and the best performance is achieved with $\epsilon_p=0.95$. From Fig.~\ref{subfig-1-2}, it can be observed that the selection of $\epsilon_p$ is a trade-off between recall rate and proposal quality. Setting it to a extremely high value leads to discarding useful proposals, while a low threshold will induct clutters to undermine the quality of clustering.

\begin{table}[t]
\setlength{\abovecaptionskip}{2mm}
\centering
\caption{Comparison with fully supervised person search models}
\begin{tabular}{c|cc|cc}
\hline\thickhline
\rowcolor{mygray}  
              &     \multicolumn{2}{c|}{PRW}    & \multicolumn{2}{c}{CUHK-SYSU}                        \\ \cline{2-5} 
\rowcolor{mygray}  
 {\multirow{-2}{*}{Method}}      & mAP                 & \multicolumn{1}{c|}{top-1}                  & mAP     & top-1   \\ 
\hline \hline  
DPM~\cite{DBLP:conf/cvpr/GirshickIDM15} &20.5  &48.3    &-  &-   \\
MGTS~\cite{DBLP:conf/eccv/ChenZOYT18} &32.6  &72.1    &83.0  &83.7   \\
RDLR~\cite{DBLP:conf/iccv/HanYZTZGS19} &42.9  &70.2    &93.0  &94.2   \\
IGPN~\cite{DBLP:conf/cvpr/DongZST20} &47.2  &87.0    &90.3  &91.4   \\
TCTS~\cite{DBLP:conf/cvpr/WangMCSC20} &46.8  &87.5    &93.9  &95.1   \\ \hline
OIM~\cite{DBLP:conf/cvpr/XiaoLWLW17} &21.3  &49.9    &75.5  &78.7   \\
IAN~\cite{DBLP:journals/pr/XiaoXTHWF19} &23.0  &61.9    &76.3  &80.1   \\
NPSM~\cite{DBLP:conf/iccv/LiuFJKZQJY17} &24.2  &53.1    &77.9  &81.2   \\
CTXGraph~\cite{DBLP:conf/cvpr/YanZNZXY19} &33.4  &73.6    &84.1  &86.5   \\
QEEPS~\cite{DBLP:conf/cvpr/MunjalATG19} &37.1  &76.7    &88.9  &89.1   \\
HOIM~\cite{DBLP:conf/aaai/ChenZO0S20} &39.8  &80.4    &89.7  &90.8   \\
BINet~\cite{DBLP:conf/cvpr/DongZST20a} &45.3  &81.7    &90.0  &90.7   \\
NAE~\cite{DBLP:conf/cvpr/ChenZYS20} &44.0  &81.1    &92.1  &92.9   \\
AlignPS~\cite{DBLP:conf/cvpr/YanLQBL00021} &45.9  &81.9    &93.1  &93.4   \\ 
SeqNet~\cite{DBLP:conf/aaai/LiM21} &46.7  &83.4    &93.8  &94.6   \\ \hline
DAPS (ours)   & \textbf{34.7} & \textbf{80.6}     & \textbf{77.6} & \textbf{79.6}    \\ \hline
\end{tabular}
\label{tab:FSPS}
\end{table}

\begin{table}[t]
\setlength{\abovecaptionskip}{2mm}
\centering
\caption{Comparison with weakly supervised person search models. * denotes training R-SiamNet together with both of CUHK-SYSU and PRW.}
\begin{tabular}{c|cc|cc}
\hline\thickhline
\rowcolor{mygray}  
              &     \multicolumn{2}{c|}{PRW}    & \multicolumn{2}{c}{CUHK-SYSU}                        \\ \cline{2-5} 
\rowcolor{mygray}  
 {\multirow{-2}{*}{Method}}      & mAP                 & \multicolumn{1}{c|}{top-1}                  & mAP     & top-1   \\ 
\hline \hline  
CGPS~\cite{yan2022exploring} &16.2  &68.0    &80.0  &82.3   \\
R-SiamNet~\cite{han2021weakly} &21.4  &75.2    &86.0  &87.1   \\
R-SiamNet*~\cite{han2021weakly} &23.5  &76.0    &86.2  &87.6   \\ \hline
DAPS (ours)   & \textbf{34.7} & \textbf{80.6}     & \textbf{77.6} & \textbf{79.6}    \\ \hline
\end{tabular}
\label{tab:WSPS}
\end{table}

\subsection{Comparison with State-of-the-Art Methods}

Since no existing person search methods with such domain adaptation settings can be directly compared, we further compare DAPS with fully supervised methods in Table~\ref{tab:FSPS}, including both of the two-step methods and one-step ones. It is surprising that our framework even surpasses some supervised methods. For example, DAPS outperforms MGTS~\cite{DBLP:conf/eccv/ChenZOYT18}, OIM~\cite{DBLP:conf/cvpr/XiaoLWLW17}, IAN~\cite{DBLP:journals/pr/XiaoXTHWF19}, NPSM~\cite{DBLP:conf/iccv/LiuFJKZQJY17} and CTXGraph~\cite{DBLP:conf/cvpr/YanZNZXY19} on PRW. The comparison with the state-of-the-art fully supervised methods indicate that there exists a large performance gap, and we hope our work will encourage more explorations for this setting. Moreover, to make measure the theoretical upper limit of DAPS setting, we train some state-of-the-art method with both datasets in a supervised manner, and more details are described in the supplementary material.

The comparisons with existing weakly supervised methods are shown in Table~\ref{tab:WSPS}, and we also present the results of training R-SiamNet with both datasets in the weakly supervised manner. When evaluated on the PRW dataset, DAPS outperforms all existing weakly supervised methods by a significant margin. For the CUHK-SYSU dataset, DAPS still underperforms the state-of-the-art weakly supervised models, which is mainly caused by the limitation brought by detection capabilities. As mentioned in Sec.~\ref{dataset}, the images and identities in PRW are prominently fewer than those in CUHK-SYSU, and this further leads to the poor detection performance of adopting CUHK-SYSU as target domain.
\label{sect:figures}

\begin{figure}[t]
    \setlength{\abovecaptionskip}{8pt}
    \centering
    \includegraphics[width=\linewidth]{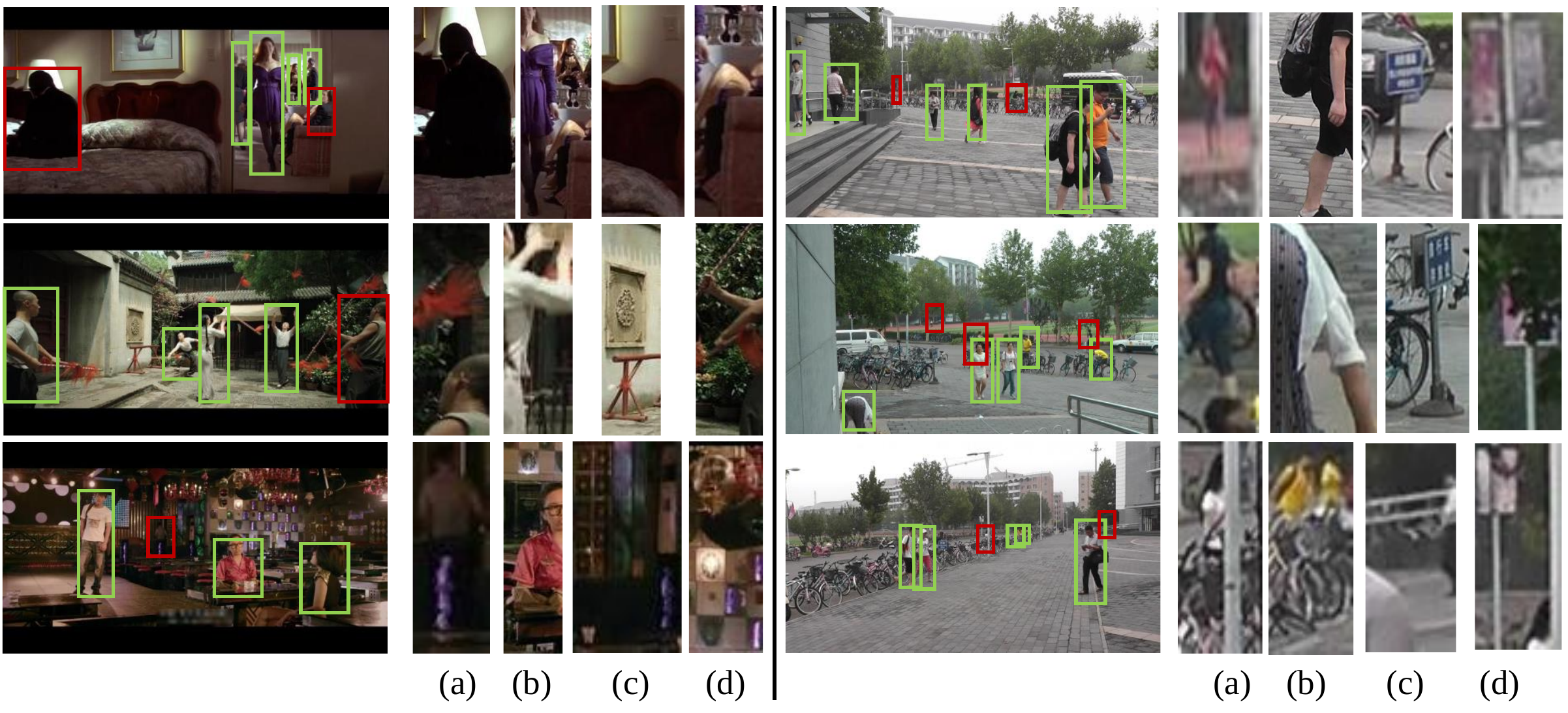}
    \caption{Visualization of some hard cases, the green bounding boxes denote the qualified proposals, while the red ones denote the undetected persons. The crops of the hybrid hard cases are presented on the right of the images.}
    \label{fig:hardcase}
\end{figure}

\subsection{Qualitative Results}
To better illustrate the distributions of our hybrid hard cases, we visualize some qualitative results from both datasets in Fig.~\ref{fig:hardcase}. As is observed, the hybrid hard cases consist of undetected persons (column a), highly overlapped human crops (column b) and background clutters(column c,d). These qualitative results demonstrate the diversity of our hybrid hard cases, and validate the rationality of adding such cases to the memory bank.

\section{Conclusions}
In this paper, we introduce a novel Domain Adaptive Person Search setting, where neither bounding boxes nor identity labels for target domain are required. Based on this new setting, we propose a strong baseline framework by investigating domain alignment and taking advantage of unlabeled target domain data. Extensive results on two large-scale benchmarks demonstrate the promising performance our framework achieves and the effectiveness of designed modules. We hope this work will encourage more exploration in this direction.

\subsubsection{Acknowledgment}
This work was supported by Shanghai Municipal Science and Technology Major Project (2021SHZDZX0102), CAAI-Huawei MindSpore Open Fund.

\clearpage
%
%
\bibliographystyle{splncs04}
\bibliography{egbib}
\end{document}